\def\year{2022}\relax
\documentclass[letterpaper]{article} 
\usepackage{aaai22}  
\usepackage{times}  
\usepackage{helvet}  
\usepackage{courier}  
\usepackage[hyphens]{url}  
\usepackage{graphicx} 
\usepackage{natbib}  
\usepackage{caption} 
\DeclareCaptionStyle{ruled}{labelfont=normalfont,labelsep=colon,strut=off} 
\usepackage{algorithm}
\usepackage{color}
\usepackage{algorithmic}
\usepackage{subfigure}
\usepackage{amssymb,amsmath,amsthm}
\newtheorem{theorem}{Theorem}

\usepackage{newfloat}
\usepackage{listings}
\floatstyle{ruled}
\newfloat{listing}{tb}{lst}{}
\floatname{listing}{Listing}

\usepackage{url}
\usepackage{bm}
\usepackage[american]{babel}
\usepackage{microtype}
\newcommand{\T}{{\top}}

\newcommand{\tr}{{\rm Tr}}

\newcommand{\eg}{{e.g.}}
\newcommand{\ie}{{i.e.}}

\renewcommand{\geq}{\geqslant}
\renewcommand{\leq}{\leqslant}
\title{Stationary Diffusion State Neural Estimation for Multiview Clustering}
\author{Chenghua Liu, Zhuolin Liao, Yixuan Ma, Kun Zhan$^\star$}
\affiliations{School of Information Science and Engineering, Lanzhou University\\
\{liuchh20,liaozhl20,mayx2021,kzhan\}@lzu.edu.cn\\
\url{https://github.com/kunzhan/SDSNE}
}

\begin{document}
\maketitle
\begin{abstract}
Although many graph-based clustering methods attempt to model the stationary diffusion state in their objectives, their performance limits to using a predefined graph. We argue that the estimation of the stationary diffusion state can be achieved by gradient descent over neural networks. We specifically design the Stationary Diffusion State Neural Estimation (SDSNE) to exploit multiview structural graph information for co-supervised learning. We explore how to design a graph neural network specially for unsupervised multiview learning and integrate multiple graphs into a unified consensus graph by a shared self-attentional module. The view-shared self-attentional module utilizes the graph structure to learn a view-consistent global graph. Meanwhile, instead of using auto-encoder in most unsupervised learning graph neural networks, SDSNE uses a co-supervised strategy with structure information to supervise the model learning. The co-supervised strategy as the loss function guides SDSNE in achieving the stationary state. With the help of the loss and the self-attentional module, we learn to obtain a graph in which nodes in each connected component fully connect by the same weight. Experiments on several multiview datasets demonstrate effectiveness of SDSNE in terms of six clustering evaluation metrics.
\end{abstract}
\section{Introduction}
Feature diversity is ubiquitous and we live in a world composed of a large amount of multiview content.
Multiple views refer to different features of the same instance~\cite{blum1998combining}.
Since multiview features are highly relevant, more and more artificial intelligence tasks involve the processing of multiview data.
Our goal is to leverage multiview data to derive clustering algorithms.

Multiview clustering carries out joint feature learning and co-view relationship modeling, aiming to exploit the correlation of different views effectively.
Since the essential consensus structure coexists in multiview features, using the structured graphs of different views for unsupervised multiview feature learning is able to optimize clustering performance.
Combining structural information from different views can achieve better efficient performance than clustering of any single view.

\begin{figure}[!t]
  \centering
  \includegraphics[width=0.46\textwidth]{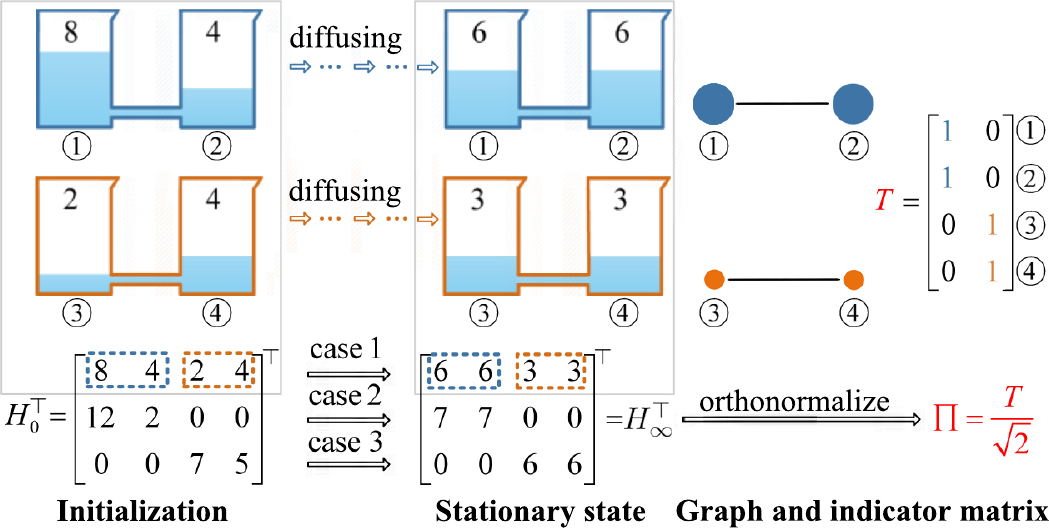}\\
\caption{A diffusion example.
A cup denotes a node in the graph and the color of cups denotes the class label. Cups have different initial water volume. After a long time, connected cups with the same color have the same amount of water, which is stationary. All of the different stationary states can be categorized into three cases, \ie, case 1 is that four cups have water as shown; case 2 is blue cups have water while no water in oranges; and case 3 is no water in blues while oranges have water, so we show three cases for example. If all of the different stationary states are orthonormalized to a stationary matrix $\Pi$, then $\Pi$ has the semantic meaning to the indicator matrix $T$. Then, we argue that the estimation of $\Pi$ can be achieved by gradient descent over neural networks.}\label{motivation}
\end{figure}

Since discriminative modeling is mostly used when exploring the internal structure of data, the structural relationship between data points is usually expressed in the form of a graph.
With predefined graphs, co-learning spectral clustering algorithms are applied firstly to unsupervised multiview learning \cite{kumar2011cot,kumar2011co} due to the well-defined mathematical framework of spectral clustering (SC)~\cite{ng2002spectral}.
Later, a series of SC or subspace learning-based methods are applied to multiview clustering by exploiting graph information~\cite{xia2014robust,li2015large,gao2015multi,zhang2018generalized}.
Although these graph-based multiview clustering methods have made great progress by integrating information from different views, their unified consensus representation is usually obtained by a simple weighted-sum approach.
We design a neural estimator that synergistically uses multiple graphs to obtain the consensus graph by exploring the internal structure of multiview data.
In this paper, we learn to obtain a consensus graph and the clustering results can be obtained without performing the eigen decomposition.

How to specifically design a graph neural network for unsupervised multiview learning is a challenging problem.
We argue that the estimation of the stationary state as described in Fig.~\ref{motivation} can be achieved by gradient descent over neural networks. In Fig.~\ref{motivation}, the stationary state is given by $\Pi$ and the values in $\Pi$ means the blue cups belong to cluster 1 and the oranges belong to cluster 2, \ie, in the graph of  Fig.~\ref{motivation}, the blues connect with blues while oranges connect to oranges, which is an ideal structure since each node-connected component belongs to one cluster. We exploit multiple graphs to learn such an ideal structural graph under the motivation of the stationary diffusion state neural estimation (SDSNE). If a multiview system becomes stationary, our intuition is that it needs to share an intrinsical structural graph. With this intuition, we use a shared self-attentional module to model a neural layer. In comparison with graph convolutional network (GCN)~\cite{kipf2016semi}, a step of the diffusion pattern is similar to the layer of GCN, but GCN lacks effective supervision to render it stationary at the ideal state. For designing a good supervision, we use the multiple graphs in a unsupervised manner via adding a co-supervised loss.

For improving the multiview clustering performance, we design SDSNE under considering three observations: (1) An intrinsical consensus graph sharing between views; (2) the unified global feature is the supervision of each single-view feature and each single-view is also the supervision of the global feature; and (3) for achieving the stationary state, the neural estimated state transition probability matrix is constructed by a learned graph. Inspired by observation 1, the parameter matrix of SDSNE models a graph and the multiview features share the same parameter matrix. From observation 2, we design a co-supervised loss function. The co-supervised loss of SDSNE not only renders the diffusion to become stationary but also makes different views consistent. Observation 3 makes SDSNE different from auto-encoder, \ie, we do not use a reconstructed loss like the auto-encoder for features or graphs. The terms of the co-supervised loss function are learned by SDSNE.

The main contributions of this paper are listed as follows:
\begin{itemize}
\item We introduce the Stationary Diffusion State Neural Estimation (SDSNE) for multiview clustering, which is trainable via back-propagation.
\item We show that the multiview clustering utility of this estimator is derived from the shared parameter between views and the unified global feature. 
\item We design a co-supervised loss to guide SDSNE in achieving the stationary state. The co-supervision of the global feature and each single view renders them achieve the stationary state.
\end{itemize}
Thus, SDSNE for multiview clustering can learn a good graph representation, and we use the graph representation to obtain the clustering label. Extensive experiments on real-world datasets are conducted to validate the superiority of SDSNE in terms of different clustering evaluation metrics.
\section{Related Works}
Superficially, the objective of SDSNE is similar to SC~\cite{ShijianboCVPR,ng2002spectral}. SDSNE uses the learned graph in its objective function but SC exploits a predefined graph. SC-based methods are very difficult to avoid the eigen decomposition of a Laplacian matrix constructed by the raw feature or the embedding. SDSNE achieves its objective by utilizing the learned multiview structural graphs synergistically. A synergism co-supervised loss function is designed to render each single view stationary. In existing traditional methods, clustering with adaptive neighbors  (CAN)~\cite{nie2014clustering} also learns to obtain exact $k$ connected components in the graph ($k$ is the cluster number.). Inspired by CAN, Zhan~\cite{zhancyb2018,Zhan8052206} designed MVGL and MCGC to obtain a graph with $k$ connected components for multiview clustering. The objective of MVGL or MCGC is divided into subproblems and is solved alternately. In MVGL and MCGC, the number of components is determined by the number of multiplicity of 0 as an eigenvalue of the Laplacian matrix. Thus, they need to perform the eigen decomposition in each iteration, which results in high complexity. Although SpectralNet~\cite{shaham2018spectralnet} and MvSCN~\cite{huang2019multi} use the concept of spectral clustering in their titles, but the representation learning of them is mainly derived from Siamese networks~\cite{hadsell2006dimensionality,LeCun1467314} or DEC~\cite{xie2016unsupervised}. The core representation learning of SpectralNet and MvSCN do not use the objective of SC. We build a neural estimator to achieve such objective in an end-to-end way effectively.

For the multiview clustering, O2MAC~\cite{fan2020one2multi} employs GCN for attributed multiview graph clustering. The graph auto-encoder of O2MAC exploits a reconstruction loss between input graphs and decoded graphs. Besides the loss, O2MAC uses a self-supervised loss. In the beginning, GCN was used for semi-supervised clustering~\cite{kipf2016semi}. If GCN has a good prediction on unlabelled data, its predicted labels are similar to the clustering indicator matrix. Since the GCN layer is quite similar to a step of the diffusion process, GCN can be regarded as using gradient descent over GCN layer for obtaining the stationary state.

Cross-diffusion process (CDP)~\cite{CVPR6248029} is inspired by co-training~\cite{blum1998combining} algorithm. CDP is applied to biomedical research for clustering cancer subtypes~\cite{wang2014similarity} and the multiview features are extracted for each patient, \ie, DNA methylation, mRNA expression, and microRNA (miRNA) expression~\cite{wang2014similarity}. Variants of CDP are widely applied to image retrieval~\cite{bai2017regularized,bai2017ensemble}. CDP has a good explanation in~\cite{bai2017regularized,bai2017ensemble}. Based on~\cite{bai2017regularized,bai2017ensemble}, CGD~\cite{tang2020cgd} uses cross-view diffusion on multiple graphs and the final clustering result is obtained by SC.

SDSNE for multiview clustering is a unified end-to-end neural network. The unique purpose of SDSNE is to estimate the stationary state. SDSNE learns to obtain a graph: nodes in each connected component connect with the same weight. With the representation it can obtain clustering with the $k$-means clustering algorithm. Although SDSNE, SC, CAN, and GCN have a similar objective, SDSNE models the stationary state directly and explores multiview clustering under the co-supervision of multiple graphs.
\section{Method}
\subsection{Denotation}
Let $\mathcal{X}=\{X^{(1)},X^{(2)},\ldots,X^{(n_v)}\}$ be a multiview dataset with $n_v$ different views.
Feature matrix is denoted by $X^{(v)}\in\mathbb{R}^{n\times d_v}$ where $n$ is the number of data points and $d_v$ denotes the dimension of the $v$-th view. We suppose that $n$ instances belong to $k$ categories. $I\in\mathbb{R}^{n\times n}$ is an identity matrix.
We build affinity graphs $A^{(v)}=[a^{(v)}_{ij}]$ for each view with the Gaussian kernel.
We model to obtain a unified graph $H=f(A^{(v)},\forall\,v|W)$ for multiview clustering, where $f(\cdot|W)$ denotes a neural network with parameter $W$.
\subsection{Stationary Diffusion State}
An undirected graph is regarded as a Markov chain~\cite{page1999pagerank}.
Graph diffusion process usually starts from a predefined affinity graph $A=[a_{ij}]\in\mathbb{R}^{n\times n},\forall\,ij,a_{ij}\geq0\,$~\cite{zhou2007spectral}. An element $a_{ij}$ denotes a pairwise affinity between nodes $i$ and $j$.
The Markov transition matrix $P$ can be deduced from $A$.
In this paper, we define $P=[p_{ij}],\forall\,ij,p_{ij}=p_{ji}\geq0,\sum_{i}p_{ij}=1\,$, and the graph diffusion process is given by
\begin{equation}\label{transition}
\bm h\leftarrow P\bm h
\end{equation}
where $\bm h$ denotes the state of nodes in the graph.

\begin{theorem}\label{the1}
The number $k$ of connected components of the graph is equal to the multiplicity of $1$ as an eigenvalue of $P\,$.
\end{theorem}
\begin{proof}
If the multiplicity of $1$ is $k$,
their corresponding orthonormal eigenvectors are $\bm{\pi}_1$, $\bm{\pi}_2,\ldots$, and $\bm{\pi}_k$\,. For each $\bm{\pi}$, we have $\bm\pi^\T \bm\pi-\bm\pi^\T P\bm\pi=0$ from $P\bm\pi= 1\bm\pi$,
then, $\bm\pi^\T \bm\pi-\bm\pi^\T P\bm\pi=\frac12\sum_{i,j=1}^np_{ij}(\pi_i-\pi_j)^2=0$ if and only if $\pi_i=\pi_j$ is constant on each connected component.
\end{proof}
\begin{theorem}\label{the2}
The following statements are equivalent for the Markov chain determined by a transition probability matrix $P$
\begin{enumerate}
  \item The Markov chain is stationary at the state of $\bm\pi$.
  \item $\bm\pi=P\bm\pi$\,.
  \item $\sum_{i,j=1}^np_{ij}(\pi_i-\pi_j)^2=0$\,.
\end{enumerate}
\end{theorem}
\begin{proof}
If $\bm h$ is unknown, the solution of the equation $\bm h=P\bm h$ is an eigenvector corresponding to the eigenvalue $1$ of $P$  or is obtained by the Gauss-Seidel method. With the Gauss-Seidel method, Eq.~\eqref{transition} iterates until $\bm h$ does not change, which means that an initial state without loss of generality is able to render it stationary at the state of $\bm\pi$, \ie, \textit{1} and \textit{2} are equivalents. From the proof of Theorem~\ref{the1}, we find that \textit{2} and \textit{3} are equivalent.
\end{proof}
\begin{theorem}\label{the3}
The $k$ eigenvectors corresponding to eigenvalue $1$ of $P$ constructs the matrix $\Pi=[\bm\pi_i]$, and $H=[\bm h_i]$ is constrained by $H^\T H=I$\,. Then, the inequality
$0=\sum_{i=1}^k{\bm \pi}_i^\T(I-P){\bm \pi}_i=\min_{H^\T H=I}\tr(H^\T(I-P)H)\leq\sum_{i=1}^k\bm h_i^\T(I-P)\bm h_i$ holds.
\end{theorem}
\begin{proof}
See~\citeauthor{Zhan8052206}~\citeyear{Zhan8052206}.
\end{proof}

From Theorem~\ref{the3}, we have
\begin{equation}
\tr(\Pi^\T(I-P)\Pi)\leq\tr(H^\T(I-P)H)
\end{equation}
and we argue that the estimation of the stationary state described in Theorem~\ref{the1} can be achieved by gradient descent over neural networks, \ie, we use $\tr(H^\T(I-P)H)$ as its loss function and define Eq.~\eqref{transition} to be its neural layer.

In comparison with GCN~\cite{kipf2016semi}, the diffusion, Eq.~\eqref{transition}, is similar to the layer of GCN. The layer of GCN can be described by
\begin{equation}
H \leftarrow \hat{A} H\Theta\,\label{gcn2}
\end{equation}
where $\Theta$ denotes the model parameter, and $\hat A$ is a normalized graph with self-loop. $\hat A$ has the same semantic meaning of the state transition probability matrix $P$.

With the guidance of some loss functions, if the final representation of GCN, $H$, tends to be a good prediction, it is similar to $\Pi$. $\Pi$ is the ideal clustering indicator, which implies that the output of Eq.~\eqref{gcn2} is directly obtained from the stationary state after learning with the gradient descent over GCN.
\subsection{SDSNE for Multiview Clustering}
In the stationary diffusion state neural estimation (SDSNE), we use multiview graphs. Given different transition matrices, $P^{(1)}$ and $P^{(2)}$, in two views, we construct a hyper transition matrix with them,
\begin{equation}
\bm{P}=P^{(1)}\otimes P^{(2)}
\end{equation}
where $\otimes$ denotes the Kronecker product. The detail of why we design such a hyper transition matrix refers to $\S$~\ref{sam}.

Then, the diffusion with $\bm{P}$ is given by
\begin{equation}
\bm{g}\leftarrow\bm{Pg}={\rm vec}\bigl(P^{(2)}S(P^{(1)})^\T\bigr)\label{bigdiff}
\end{equation}
where ${\rm vec}(\cdot)$ denotes the vectorization by stacking columns one by one and ${\rm vec}(S)=\bm{g}$\,.

If a multiview system becomes stationary, our intuition is that it needs to share an intrinsical structural graph.
From Eq.~\eqref{bigdiff}, we model the consensus graph $S$ for different views by,
\begin{equation}
S\leftarrow P^{(v)}S(P^{(u)})^\T,\forall\,u,v\in\{1,2,\ldots,n_v\}\,.\label{layer}
\end{equation}
Eq.~\eqref{layer} implies that all views share a consensus graph $S$. Note that $S$ can also be regarded as a graph and the detail of why $S$ is a graph refers to $\S$~\ref{sam}.

We use a neural network to learn directly with the gradient descent algorithm to obtain such a stationary state. Since the consensus feature is modeled by $S$ which is shared between different views, $P$ in Eq.~\eqref{layer} can be from the same view. We model the diffusion as a layer of the neural network and we share model parameter $W$ in different views,
\begin{equation}\label{Layer}
H^{(v)}\leftarrow P^{(v)}W(P^{(v)})^\T\,,\forall\,v\in\{1,2,\ldots,n_v\}\,.
\end{equation}

Then, we fuse the learned features $H^{(v)}$ to obtain a unified global feature by,
\begin{equation}
H=\alpha \sum_{v=1}^{n_v}{H^{(v)}}+(1-\alpha)I\,\label{self_loop}
\end{equation}
where $\alpha$ is a trade-off hyper parameter.

Note that $H^{(v)}$ can be also regarded as a graph and the detail of why $H^{(v)}$ is a graph refers to $\S$~\ref{sam}. With Eq.~\eqref{Layer}, we obtain different graphs $H^{(v)}$ and $H^{(v)}$ is normalized to attain $\hat P^{(v)}$ for each view.

According to Theorems~\ref{the2} and \ref{the3}, the loss function guides SDSNE in obtaining a stationary state by minimizing,
\begin{equation}\label{SDSNE}
\mathcal{L}_{\rm sds}=\sum_{v=1}^{n_v}\tr\bigl(H^\T(I-\hat{P}^{(v)})H \bigr)\,.
\end{equation}

If a hyper graph achieves the stationary state, edges in each connected component tend to have the same value in the output graph $H^{(v)}$, \ie, nodes in each component connect with each other by the same edge weight. We need a graph in which different values in different components, rather than a graph in which a component is marked by a non-zero value while others are zeros, so we add an $\ell_2$-regularization loss. The overall loss is given by,
\begin{equation}\label{overall}
\mathcal{L}=\mathcal{L}_{\rm sds} + \mu\sum_{v=1}^{n_v}\tr\bigl((H^{(v)})^\T H^{(v)}\bigr)\,.
\end{equation}

We summarize the SDSNE algorithm in Algorithm~\ref{algorithm}. We use the symmetric Laplacian matrix rather than the random walk Laplacian since the former usually has better performance~\cite{chung1997spectral,von2007tutorial,ShijianboCVPR}.
\begin{algorithm}[htbp]
\caption{SDSNE for multiview clustering.}\label{algorithm}
\begin{algorithmic}[1]
\STATE \textbf{Input}: $\mathcal{X}=\{X^{(1)},X^{(2)},\ldots,X^{(n_v)}\}$\,.
\STATE \textbf{Output}: $H$\,.
\STATE \textbf{Initialization:} $\alpha$, $\mu$, $W$, and $epoch_{\max}$\,.
\FOR{$v \in \{1,2,\dots, n_v\}$}
\STATE Construct $A^{(v)}$ by the Gaussian kernel with $X^{(v)}$\,.
\STATE Calculate the degree matrix $D^{(v)}$ of $A^{(v)}$\,.
\STATE Normalize $A^{(v)}$ by $P^{(v)}=(D^{(v)})^{-\frac12}A^{(v)}(D^{(v)})^{-\frac12}$\,.
\ENDFOR
\WHILE{$epoch\leq {epoch_{\max}}$}
    \FOR{$v \in \{1,2,\dots, n_v\}$}
        \STATE $H^{(v)}\leftarrow P^{(v)}W(P^{(v)})^\T\,.$
        \STATE Calculate the degree matrix $\hat D^{(v)}$ of $H^{(v)}$\,.
        \STATE $\hat P^{(v)}=(\hat D^{(v)})^{-\frac12}H^{(v)}(\hat D^{(v)})^{-\frac12}$\,.
    \ENDFOR
    \STATE Update $H$ by Eq.~\eqref{self_loop}\,.
    \STATE Update $\mathcal L$ by Eq.~\eqref{overall}\,.
    \STATE Update $W$ by the gradient descent algorithm.
    \IF {The loss $\mathcal L$ converges.}
        \STATE Break.
    \ENDIF
    \STATE $epoch = epoch + 1\,.$
\ENDWHILE
\end{algorithmic}
\end{algorithm}

\subsection{Analysis of SDSNE}\label{sam}
First, we give the reason why we use a hyper graph. Referring to Theorem~\ref{the3} and Eq.~\eqref{bigdiff}, the third statement in Theorem~\ref{the2} can be reached by minimizing,
\begin{equation}\label{OBJ}
\bm g^\T(\bm I-\bm P)\bm g=\frac12\sum_{i,j,k,l=1}^np^{(1)}_{ij}p^{(2)}_{kl}(s_{ik}-s_{jl})^2
\end{equation}
where $\bm I\in\mathbb{R}^{n^2\times n^2}$ is an identity matrix and $\bm g={\rm vec}(S)$. $S=[\bm s_i]$ is a symmetric similarity matrix. In the right side of Eq.~\eqref{OBJ}, if fixing $p_{kl}$, minimizing it makes $\bm s_i$ and $\bm s_j$ similar when $p_{ij}$ is large; if fixing $p_{ij}$, $\bm s_k$ and $\bm s_l$ are similar when $p_{kl}$ is large; if both $p_{ij}$ and $p_{kl}$ are large, the affinity $s_{ik}$ tends to the same value of $s_{jl}$. Thus, by minimizing $\sum_{i,j,k,l=1}^n\hat p^{(1)}_{ij}\hat p^{(2)}_{kl}(h_{ik}-h_{jl})^2$, SDSNE achieves a graph $H^{(v)}$ in which nodes in each connected component connects with each other by the same edge weight. Since minimizing $\sum_{i,j,k,l=1}^n\hat p^{(1)}_{ij}\hat p^{(2)}_{kl}(h_{ik}-h_{jl})^2$ is equal to minimizing $\hat p^{(v)}_{kl}\sum_{i,j=1}^n\hat p^{(v)}_{ij}(h_{ik}-h_{jl})^2=p^{(v)}_{kl}\tr\bigl(H^\T(I-\hat{P}^{(v)})H \bigr)$, we use Eq.~\eqref{SDSNE} as the loss function to render each view stationary.

Second, we answer why $S$ or $H^{(v)}$ is regarded as a graph. Since Eq.~\eqref{Layer} is a self-attentional module, we regards $S$ or $H^{(v)}$ as a graph. We suppose that using $P^{(v)}$ learns a query $Q^{(v)}$ and a key $K^{(v)}$,
\begin{eqnarray}
Q^{(v)}&=&P^{(v)}W_1\,, \\
K^{(v)}&=&P^{(v)}W_2
\end{eqnarray}
where $W_1$ and $W_2$ are two learnable weight matrices. We use $Q^{(v)}$ and $K^{(v)}$ to construct a new self-attentional graph, the attentional coefficient matrix can be given by Eq.~\eqref{Layer}, \ie,
\begin{equation}
H^{(v)}\leftarrow Q^{(v)}(K^{(v)})^\T=P^{(v)}W(P^{(v)})^\T\,.
\end{equation}


Third, we analyze the reason why we use $\hat P^{(v)}$ rather than $P^{(v)}$ in the loss. We suppose that the stationary $H^{(v)}_\infty=\Pi^{(v)}$ achieves, then we have $\Pi^{(v)}=\hat P^{(v)}H_0^{(v)}$, where $H_0^{(v)}$ is the initial state. This stationary state also can be derived from $\Pi^{(v)}=P^{(v)}P^{(v)} \cdots P^{(v)}H_0^{(v)}$ without loss of generality. It means that SDSNE models $W=P^{(v)}P^{(v)} \cdots P^{(v)}H_0^{(v)}$ directly by gradient descent algorithm and $W$ is shared between views. If we use the fixed $P^{(v)}$, it guides SDSNE in staying at the first step of diffusion.

Fourth, we analyze the effect of the self-loop. We add a weighted self-loop in Eq.~\eqref{self_loop}. The loss, Eq.~\eqref{SDSNE}, also is a structural view-consistent loss, and the self-loop in $H$ guides $\hat P^{(v)}$ in learning a self-loop. As shown in Fig.~\ref{motivation}, if the transition probability to others is 1 and the probability to itself is 0, all water moves to others unreservedly at the first step of diffusion, and all will return at the next step. To avoid such oscillations, we add a weighted self-loop in Eq.~\eqref{self_loop}\,.
\subsection{Convergence and Complexity Analysis}
Since \eqref{overall} is convex, i.e., the Laplacian matrix is positive semi-definite~\cite{chung1997spectral}, minimizing it by the gradient descent algorithm renders SDSNE converged.

SDSNE does not perform the eigen decomposition. The complexity of the eigen decomposition is $O(n^3)$ where $n$ is the number of data points. The layer of SDSNE costs $O(n^3)$ complexity too since it needs to perform the matrix multiplication. The matrix multiplication is easier to calculate by a parallel algorithm than the eigen decomposition. Since the existing cross-diffusion methods need $O(n^4)$ to compute the objective, SDSNE has lower complexity than cross-diffusion-based methods.

\begin{table*}[!tp]\setlength{\tabcolsep}{1.5pt}
\centering
\caption{Clustering performance between \textbf{SDSNE} and other state-of-the-art methods.}
\label{with}
\centering
\begin{tabular*}{\textwidth}{@{\extracolsep{\fill}\quad}lcccccc}
\hline
~~~~~~Methods&NMI&ACC&ARI&$F_1$-score&Precision&Purity\\
\hline
\multicolumn{1}{l}{BBC Sport} \\
~~~~~~SC$_{\rm best}$
&0.022$\pm$0.005&0.360$\pm$0.003&0.005$\pm$0.003&0.386$\pm$0.002&0.241$\pm$0.002&0.362$\pm$0.001\\
~~~~~~LRR$_{\rm best}$
&0.775$\pm$0.002&0.904$\pm$0.003&0.747$\pm$0.001&0.812$\pm$0.002&0.754$\pm$0.001&0.904$\pm$0.001\\
~~~~~~MCGC
&0.112$\pm$0.000&0.421$\pm$0.000&0.049$\pm$0.000&0.401$\pm$0.000&0.258$\pm$0.000&0.444$\pm$0.000\\
~~~~~~GMC
&0.705$\pm$0.000&0.739$\pm$0.000&0.601$\pm$0.000&0.721$\pm$0.000&0.573$\pm$0.000&0.763$\pm$0.000\\
~~~~~~CGD
&0.910$\pm$0.003&0.974$\pm$0.004&0.931$\pm$0.002&0.947$\pm$0.001&0.943$\pm$0.003&0.974$\pm$0.002\\
~~~~~~O2MAC
&0.891$\pm$0.018&0.964$\pm$0.008&0.906$\pm$0.019&0.965$\pm$0.009&0.959$\pm$0.011&0.964$\pm$0.008\\
~~~~~~{\bf SDSNE}$_{\rm km}$
&0.899$\pm$0.000&0.969$\pm$0.000&0.918$\pm$0.000&0.938$\pm$0.000&0.974$\pm$0.000&0.969$\pm$0.000\\
~~~~~~{\bf SDSNE}$_{\rm sc}$
&0.948$\pm$0.000&0.985$\pm$0.000&0.958$\pm$0.000&0.968$\pm$0.000&0.991$\pm$0.000&0.985$\pm$0.000\\
\hline
\multicolumn{1}{l}{MSRC-v1}\\
~~~~~~SC$_{\rm best}$
&0.556$\pm$0.000&0.519$\pm$0.000&0.289$\pm$0.000&0.431$\pm$0.000&0.300$\pm$0.000&0.523$\pm$0.000\\
~~~~~~LRR$_{\rm best}$
&0.539$\pm$0.021&0.681$\pm$0.018&0.413$\pm$0.019&0.498$\pm$0.017&0.476$\pm$0.019&0.681$\pm$0.018\\
~~~~~~MCGC
&0.692$\pm$0.000&0.776$\pm$0.000&0.630$\pm$0.000&0.685$\pm$0.000&0.640$\pm$0.000&0.785$\pm$0.000\\
~~~~~~GMC
&0.816$\pm$0.000&0.895$\pm$0.000&0.767$\pm$0.000&0.799$\pm$0.000&0.786$\pm$0.000&0.895$\pm$0.000\\
~~~~~~CGD
&0.842$\pm$0.004&0.910$\pm$0.006&0.790$\pm$0.003&0.819$\pm$0.004&0.804$\pm$0.005&0.910$\pm$0.005\\
~~~~~~O2MAC
&0.617$\pm$0.011&0.709$\pm$0.030&0.525$\pm$0.020&0.691$\pm$0.026&0.716$\pm$0.040&0.715$\pm$0.021\\
~~~~~~{\bf SDSNE}$_{\rm km}$
&0.898$\pm$0.000&0.943$\pm$0.000&0.867$\pm$0.000&0.886$\pm$0.000&0.953$\pm$0.000&0.943$\pm$0.000\\
~~~~~~{\bf SDSNE}$_{\rm sc}$
&0.872$\pm$0.000&0.933$\pm$0.000&0.845$\pm$0.000&0.867$\pm$0.000&0.942$\pm$0.000&0.933$\pm$0.000\\
\hline
\multicolumn{1}{l}{100 Leaves}\\
~~~~~~SC$_{\rm best}$
&0.777$\pm$0.002&0.483$\pm$0.014&0.203$\pm$0.008&0.215$\pm$0.008&0.128$\pm$0.007&0.520$\pm$0.003\\
~~~~~~LRR$_{\rm best}$
&0.715$\pm$0.018&0.488$\pm$0.013&0.307$\pm$0.011&0.315$\pm$0.010&0.274$\pm$0.007&0.529$\pm$0.009\\
~~~~~~MCGC
&0.834$\pm$0.000&0.727$\pm$0.000&0.410$\pm$0.000&0.418$\pm$0.000&0.290$\pm$0.000&0.747$\pm$0.000\\
~~~~~~GMC
&0.902$\pm$0.000&0.824$\pm$0.000&0.497$\pm$0.000&0.504$\pm$0.000&0.352$\pm$0.000&0.851$\pm$0.000\\
~~~~~~CGD
&0.943$\pm$0.007&0.859$\pm$0.005&0.821$\pm$0.006&0.823$\pm$0.004&0.770$\pm$0.006&0.881$\pm$0.005\\
~~~~~~O2MAC
&0.782$\pm$0.003&0.557$\pm$0.009 &0.432$\pm$0.007&0.546$\pm$0.010&0.567$\pm$0.010&0.586$\pm$0.009\\
~~~~~~{\bf SDSNE}$_{\rm km}$
&0.979$\pm$0.000&0.962$\pm$0.000&0.934$\pm$0.000&0.935$\pm$0.000&0.965$\pm$0.000&0.966$\pm$0.000\\
~~~~~~{\bf SDSNE}$_{\rm sc}$
&0.972$\pm$0.000&0.957$\pm$0.000&0.913$\pm$0.000&0.914$\pm$0.000&0.967$\pm$0.000&0.957$\pm$0.000\\
\hline
\multicolumn{1}{l}{Three Sources}\\
~~~~~~SC$_{\rm best}$
&0.054$\pm$0.014&0.331$\pm$0.015&0.011$\pm$0.012&0.362$\pm$0.011&0.228$\pm$0.008&0.349$\pm$0.013\\
~~~~~~LRR$_{\rm best}$
&0.525$\pm$0.016&0.627$\pm$0.009&0.351$\pm$0.011&0.555$\pm$0.012&0.411$\pm$0.013&0.668$\pm$0.008\\
~~~~~~MCGC
&0.075$\pm$0.000&0.301$\pm$0.000&0.037$\pm$0.000&0.337$\pm$0.000&0.216$\pm$0.000&0.384$\pm$0.000\\
~~~~~~GMC
&0.548$\pm$0.000&0.692$\pm$0.000&0.443$\pm$0.000&0.605$\pm$0.000&0.484$\pm$0.000&0.746$\pm$0.000\\
~~~~~~CGD
&0.695$\pm$0.005&0.781$\pm$0.006&0.611$\pm$0.005&0.709$\pm$0.006&0.651$\pm$0.007&0.828$\pm$0.003\\
~~~~~~O2MAC
&0.727$\pm$0.030&0.755$\pm$0.026&0.650$\pm$0.040&0.669$\pm$0.022&0.667$\pm$0.025&0.840$\pm$0.020\\
~~~~~~{\bf SDSNE}$_{\rm km}$
&0.747$\pm$0.000&0.828$\pm$0.000&0.741$\pm$0.000&0.802$\pm$0.000&0.720$\pm$0.000&0.846$\pm$0.000\\
~~~~~~{\bf SDSNE}$_{\rm sc}$
&0.848$\pm$0.000&0.935$\pm$0.000&0.867$\pm$0.000&0.898$\pm$0.000&0.927$\pm$0.000&0.935$\pm$0.000\\
\hline
\multicolumn{1}{l}{Scene-15} \\
~~~~~~SC$_{\rm best}$
&0.384$\pm$0.014&0.377$\pm$0.013&0.208$\pm$0.001&0.272$\pm$0.014&0.234$\pm$0.014&0.404$\pm$0.014\\
~~~~~~LRR$_{\rm best}$
&0.369$\pm$0.002&0.368$\pm$0.003&0.201$\pm$0.001&0.263$\pm$0.002&0.233$\pm$0.003&0.395$\pm$0.001\\
~~~~~~MCGC
&0.142$\pm$0.000&0.179$\pm$0.000&0.054$\pm$0.000&0.170$\pm$0.000&0.096$\pm$0.000&0.186$\pm$0.000\\
~~~~~~GMC
&0.058$\pm$0.000&0.140$\pm$0.000&0.004$\pm$0.000&0.132$\pm$0.000&0.071$\pm$0.000&0.146$\pm$0.000\\
~~~~~~CGD
&0.419$\pm$0.006&0.428$\pm$0.004&0.256$\pm$0.003&0.315$\pm$0.003&0.277$\pm$0.002&0.484$\pm$0.004\\
~~~~~~O2MAC
&0.325$\pm$0.009&0.309$\pm$0.013&0.155$\pm$0.007&0.306$\pm$0.013&0.319$\pm$0.018&0.339$\pm$0.010\\
~~~~~~{\bf SDSNE}$_{\rm km}$
&0.437$\pm$0.000&0.443$\pm$0.000&0.247$\pm$0.000&0.308$\pm$0.000&0.505$\pm$0.000&0.458$\pm$0.000\\
~~~~~~{\bf SDSNE}$_{\rm sc}$
&0.438$\pm$0.000&0.436$\pm$0.000&0.263$\pm$0.000&0.325$\pm$0.000&0.426$\pm$0.000&0.485$\pm$0.000\\
\hline
\multicolumn{1}{l}{Reuters} \\
~~~~~~SC$_{\rm best}$
&0.112$\pm$0.012&0.296$\pm$0.008&0.059$\pm$0.000&0.378$\pm$0.007&0.238$\pm$0.009&0.329$\pm$0.007\\
~~~~~~LRR$_{\rm best}$
&0.206$\pm$0.006&0.397$\pm$0.003&0.064$\pm$0.005&0.324$\pm$0.004&0.240$\pm$0.005&0.294$\pm$0.005\\
~~~~~~MCGC
&0.263$\pm$0.000&0.439$\pm$0.000&0.072$\pm$0.000&0.388$\pm$0.000&0.257$\pm$0.000&0.349$\pm$0.000\\
~~~~~~GMC
&0.274$\pm$0.000&0.472$\pm$0.000&0.078$\pm$0.000&0.391$\pm$0.000&0.262$\pm$0.000&0.351$\pm$0.000\\
~~~~~~CGD
&0.287$\pm$0.005&0.492$\pm$0.004&0.082$\pm$0.003&0.422$\pm$0.003&0.279$\pm$0.003&0.367$\pm$0.003\\
~~~~~~O2MAC
&0.290$\pm$0.026&0.459$\pm$0.039&0.243$\pm$0.053&0.376$\pm$0.028&0.394$\pm$0.024&0.550$\pm$0.039\\
~~~~~~{\bf SDSNE}$_{\rm km}$
&0.388$\pm$0.000&0.516$\pm$0.000&0.210$\pm$0.000&0.457$\pm$0.000&0.491$\pm$0.000&0.581$\pm$0.000\\
~~~~~~{\bf SDSNE}$_{\rm sc}$
&0.393$\pm$0.000&0.522$\pm$0.000&0.237$\pm$0.000&0.471$\pm$0.000&0.484$\pm$0.000&0.587$\pm$0.000\\
\hline
\end{tabular*}
\end{table*}
\section{Experiments}
\subsection{Datasets}
Six benchmark datasets are used to demonstrate the effectiveness of the proposed method, including

BBC Sport\footnote{\url{http://mlg.ucd.ie/datasets/bbc.html}}: The document dataset contains 544 documents in five classes, such as athletics, cricket, football, rugby, tennis. Two different features are extracted for each document~\cite{xia2014robust}.

MSRC-v1\footnote{\url{https://www.microsoft.com/en-us/research/project/image-understanding/}}: The image dataset consists of seven classes: tree, building, airplane, cow, face, car, and bicycle. It contains 30 images in each category and each image has six views.

100 Leaves\footnote{\url{https://achieve.ics.uci.edu/ml/datasets/One-hundred+plant+species+leaves+data+set}}: The image dataset consists of 100 classes of leaves, there are 16 images in each class, and three different features are extracted: shape, margin, and texture.

Three Sources\footnote{\url{http://mlg.ucd.ie/datasets/3sources.html}}: The document dataset has 169 stories reported in BBC, Reuters, and the \textit{Guardian}. Each story was manually annotated with one of the six topical labels: business, entertainment, health, politics, sport, and technology.

Scene-15 \cite{fei2005bayesian}: It consists of 4485 images in total, which has 15 scene categories with both indoor and outdoor environments. For every image, three features, including GIST, PHOG, and LBP are extracted.

Reuters\footnote{\url{http://ama.liglab.fr/~amini/DataSets/Classification/Multiview/ReutersMutliLingualMultiView.htm}}: We use a subset of Reuters that consists of 18,758 articles in six classes and each article has five views, \ie, English, French, German, Italian, and Spanish.

\subsection{Experimental Setup}
We evaluate the performance of SDSNE on six multiview datasets.
We compare SDSNE with six state-of-the-art algorithms. The baseline can be coarsely categorized into three groups.
\begin{itemize}
\item[] The Best Single-view:
\begin{enumerate}
\item SC$_{\rm best}$ \cite{ng2002spectral} is performed for each single-view feature and we report the best.
\item LRR$_{\rm best}$ \cite{6180173} uses low-rank representation to solve the subspace clustering problem and we report the best single-view results.
\end{enumerate}
\item[] Graph-based:
\begin{enumerate}\setcounter{enumi}{2}
\item MCGC~\cite{Zhan8052206} imposes a rank constraint on the Laplacian matrix and utilizes a new disagreement cost function for regularizing graphs from different views to learn a consensus graph.
\item GMC~\cite{8662703} fuses the multiple graphs to generate a unified graph under the consideration to the view weights.
\item CGD \cite{tang2020cgd} learns a unified graph for multiview clustering via cross-view graph diffusion.
\end{enumerate}

\item[] GNN-based:
\begin{enumerate}\setcounter{enumi}{5}
\item O2MAC \cite{fan2020one2multi} assumes that there is a dominated view. Using GCN processes the dominated view feature to obtain a unique latent feature. With the latent feature, O2MAC reconstructs multiview graphs.
\end{enumerate}
\end{itemize}

\begin{figure*}[ht]
\centering
\subfigure[t-SNE of $X^{(1)}$]{\includegraphics[width=1.52in]{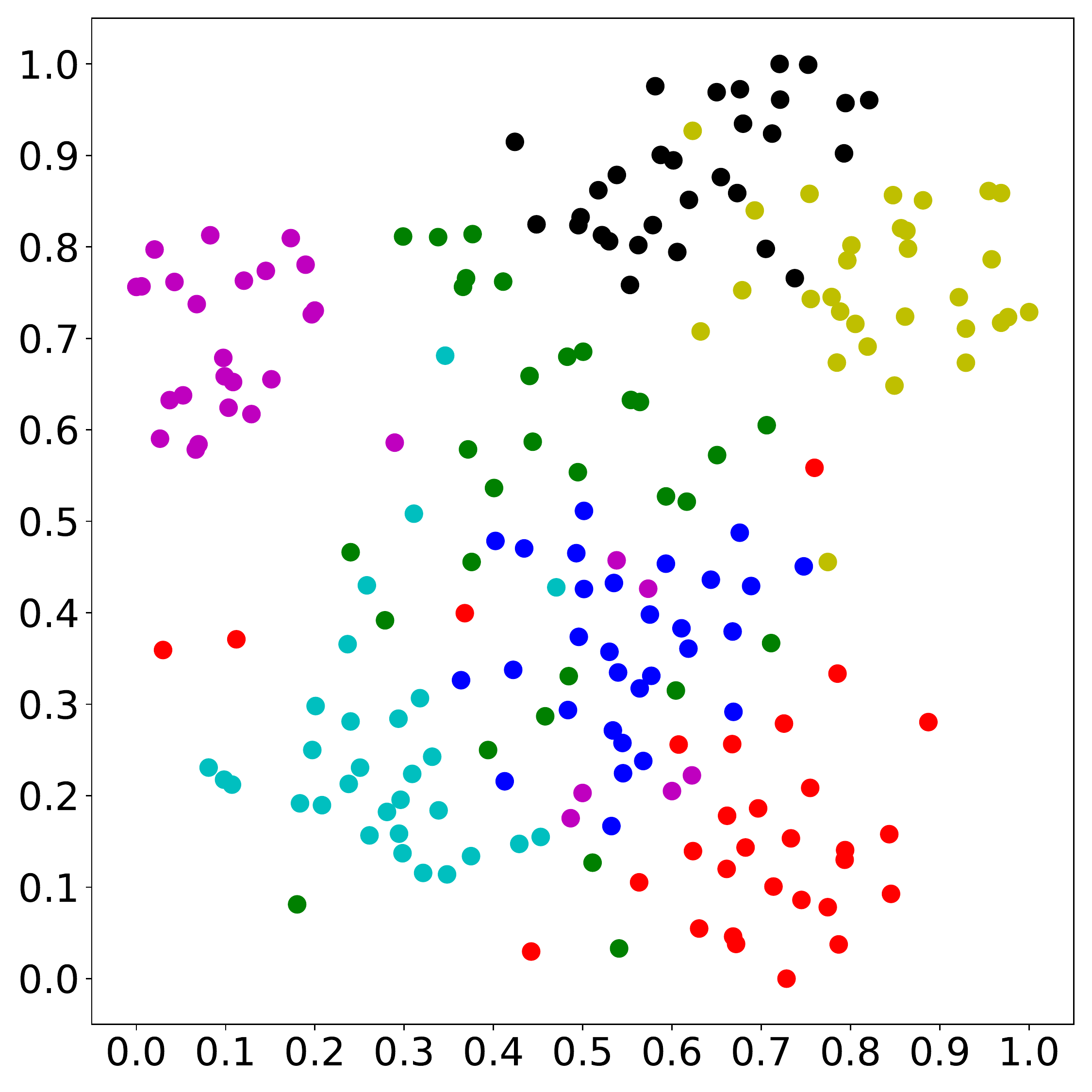}}~~~~~~~~~~
\subfigure[t-SNE of $X^{(2)}$]{\includegraphics[width=1.52in]{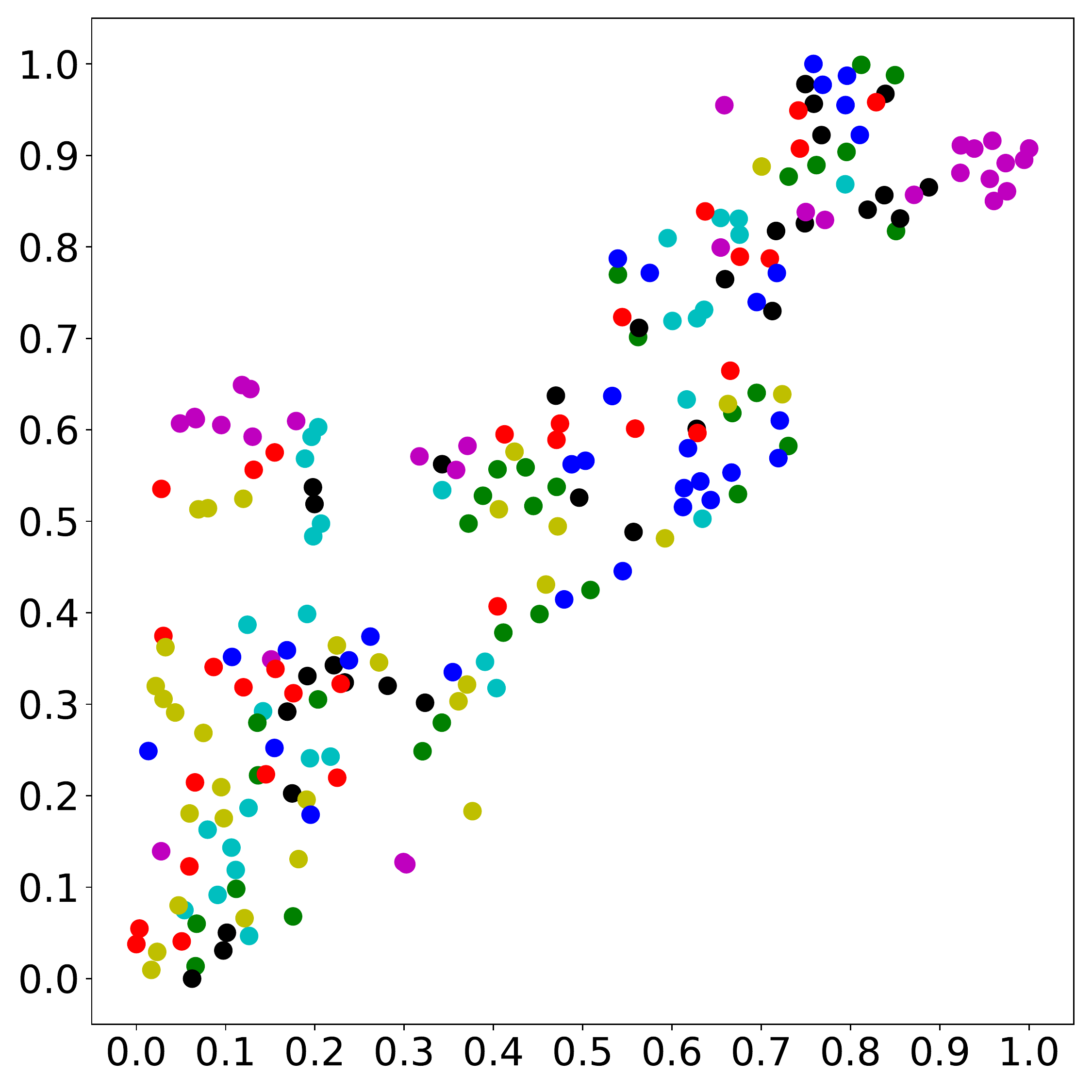}}~~~~~~~~~~
\subfigure[t-SNE of $X^{(3)}$]{\includegraphics[width=1.52in]{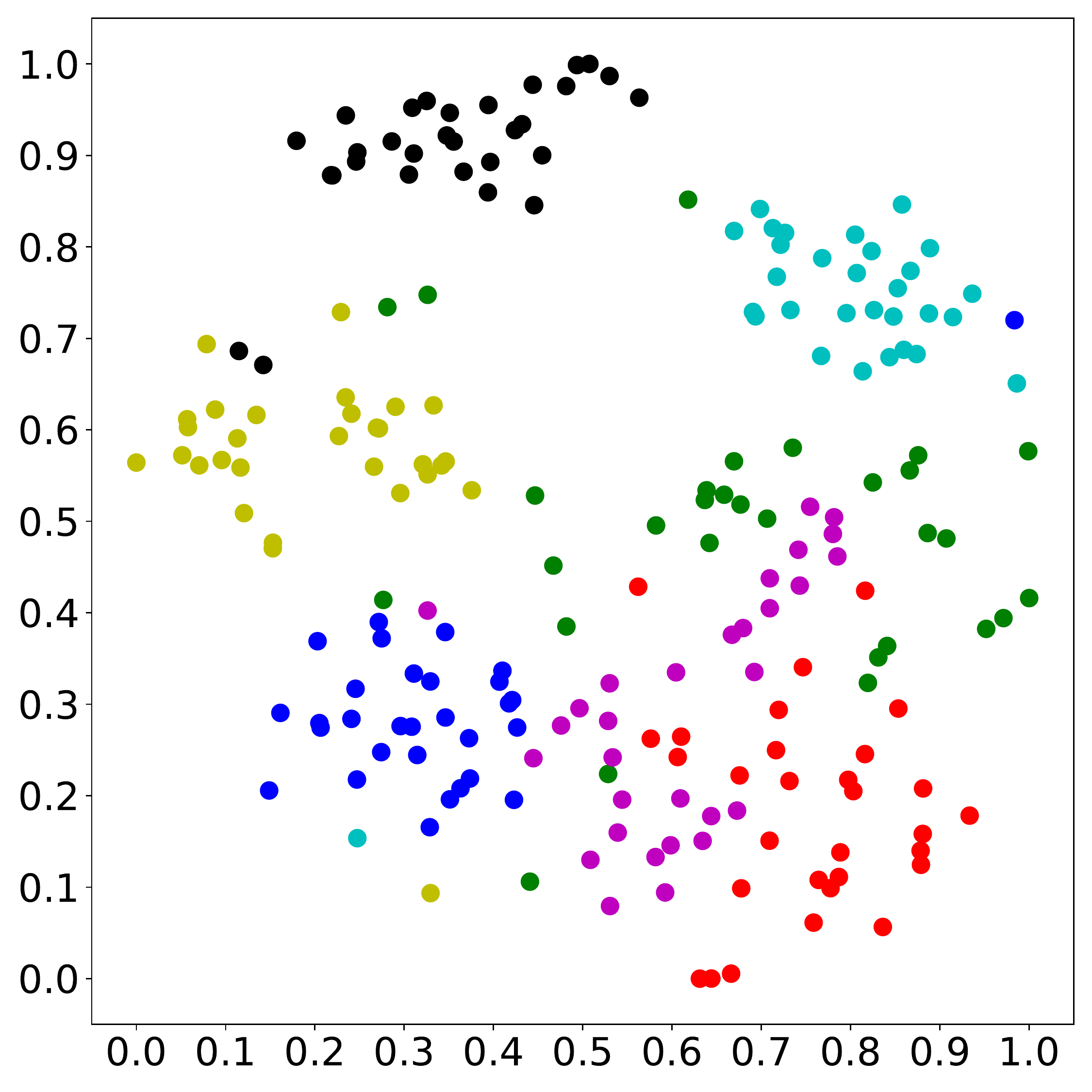}}~~~~~~~~~~
\subfigure[t-SNE of $X^{(4)}$]{\includegraphics[width=1.52in]{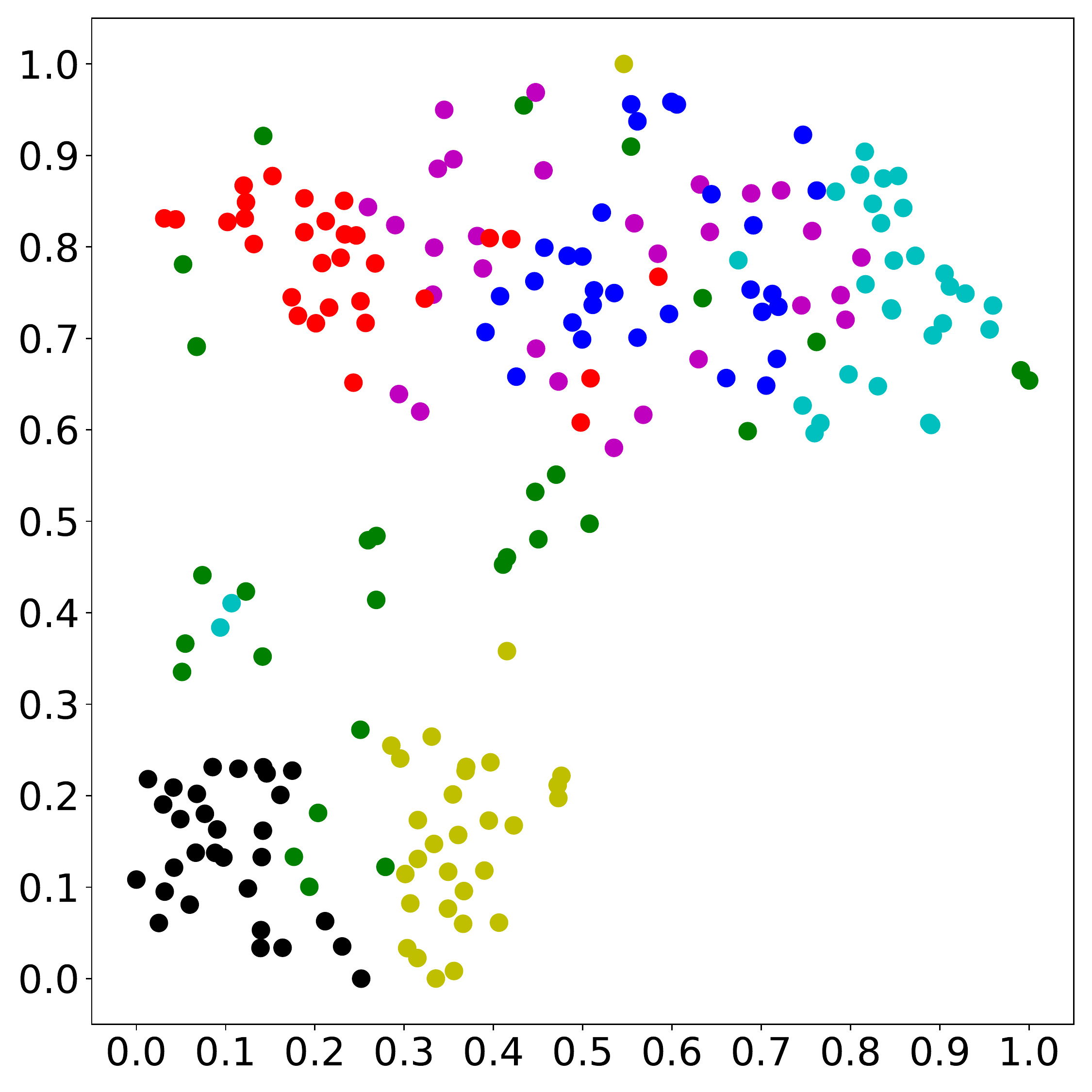}}\\
\subfigure[t-SNE of $X^{(5)}$]{\includegraphics[width=1.52in]{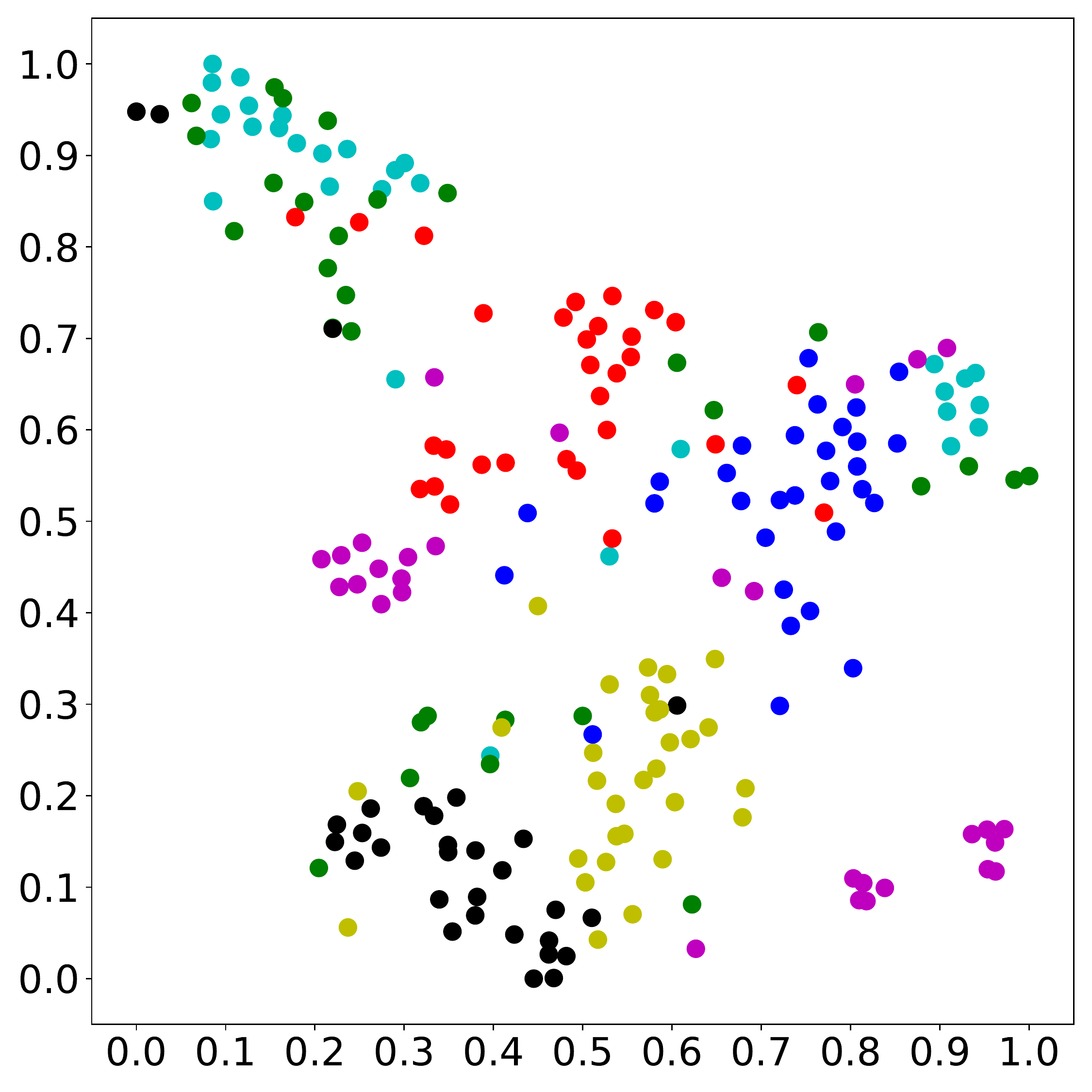}}~~~~~~~~~~
\subfigure[t-SNE of $X^{(6)}$]{\includegraphics[width=1.52in]{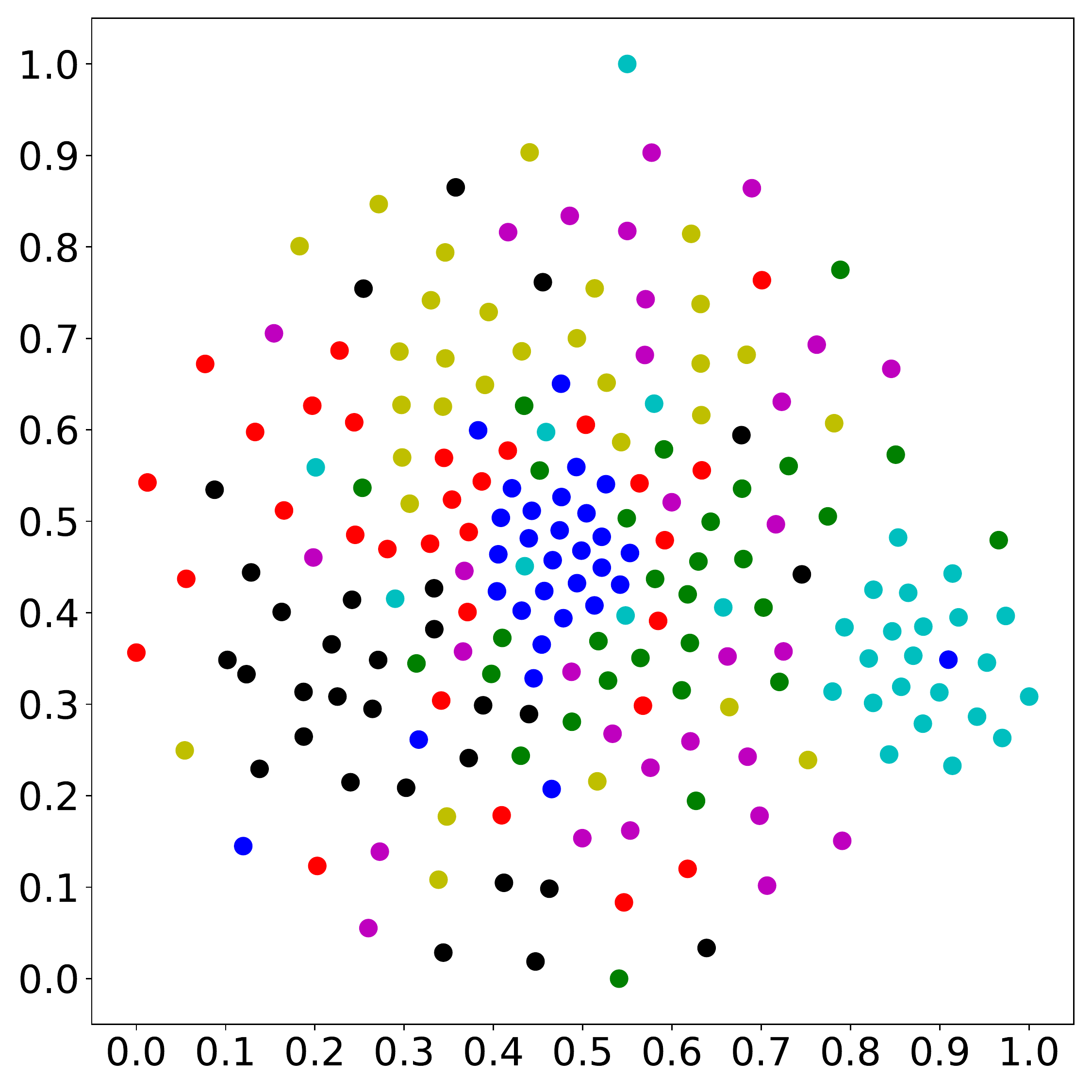}}~~~~~~~~~~
\subfigure[t-SNE of the learned $H$]{\includegraphics[width=1.52in]{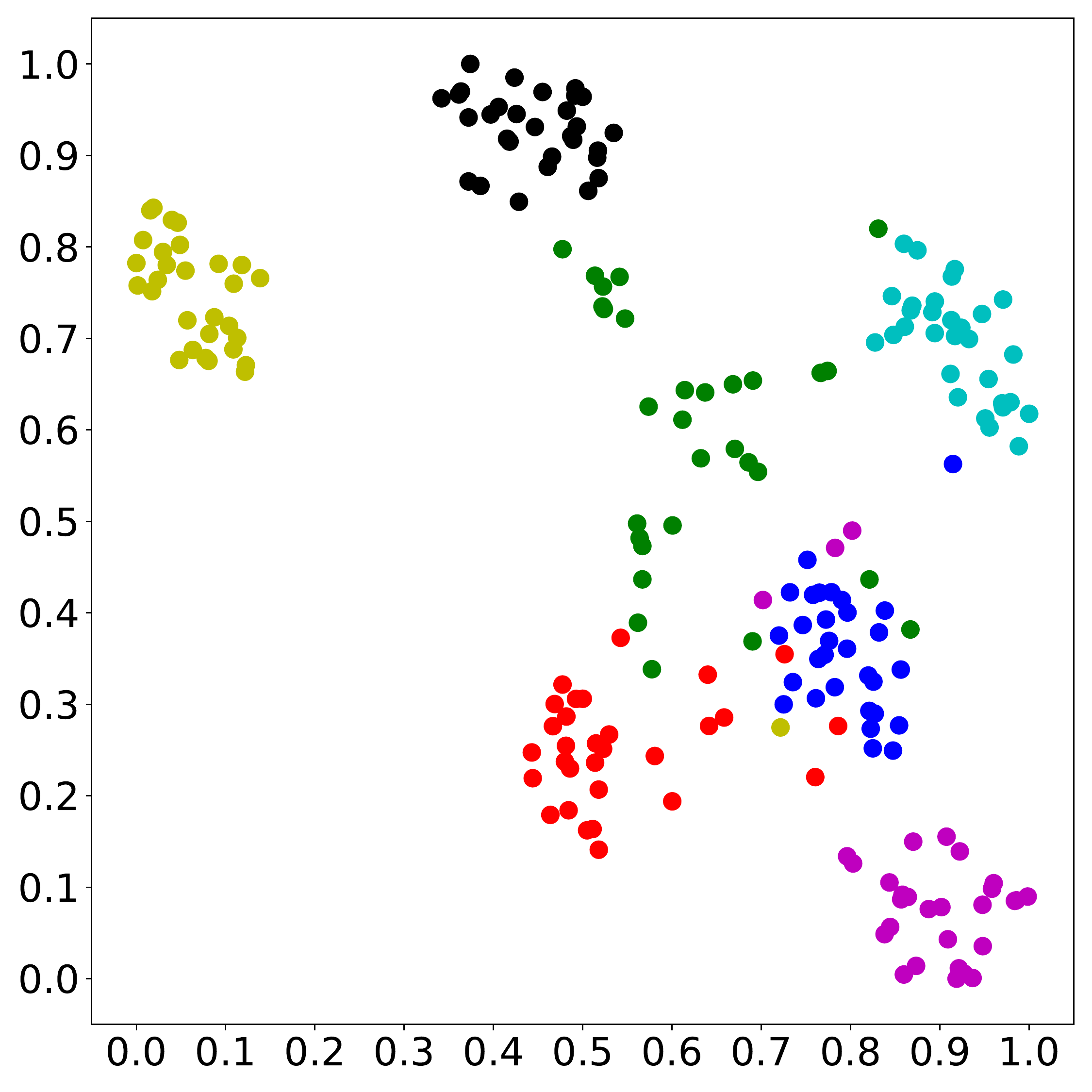}}~~~~~~~~~~
\subfigure[Heat map of $H$]{\includegraphics[width=1.52in]{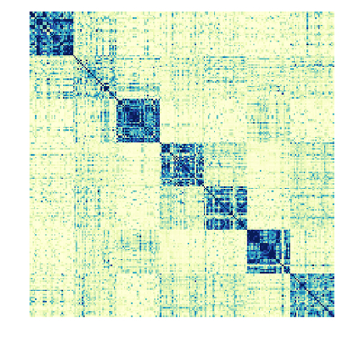}}
\caption{The visualization of the raw features and the learned $H$ of SDSNE on MSRC-v1.}\label{filter}
\end{figure*}

For a fair comparison, we run each method 10 times and report the mean of performance as well as the standard deviation. For SDSNE, we set the seed of the pseudo-random generator as in GCN~\cite{kipf2016semi} to eliminate the fluctuation of clustering results. The learning rate is set to be $10^{-4}$ or $10^{-5}$. Without loss of generality, the Gaussian kernel function with Euclidean distance is used to generate initial view-specific graphs, and the $\sigma$ in the Gaussian kernel function is set to 0.5.
In experiments, we employ six widely used metrics to measure the clustering performance: normalized mutual information (NMI), clustering accuracy (ACC), adjusted rand index (ARI), $F_1$-score, precision, and purity. Note that a higher value indicates better performance for the six metrics. During training, we use an early stop strategy with the patience of 10 and stop training when the loss function drops dramatically. We perform the $k$-means clustering and SC to obtain the clustering results.
\subsection{Experimental Results}
Clustering performance is summarized in Table~\ref{with}. SDSNE$_{\rm km}$ means that we perform the $k$-means clustering on $H$, and SDSNE$_{\rm sc}$ means that SC is performed on $H$.

In Table~\ref{with}, we obtain the following observations: (1) On all six multiview datasets, results of SDSNE$_{\rm sc}$ are higher than state-of-the-art methods. It implies that SDSNE learns the shared information well between multiple views to improve the clustering performance. (2) In most cases, SDSNE outperforms other state-of-the-art methods on both large multiview datasets, \eg, Reuters, and small datasets, \eg, Three Sources. Some methods, \eg, O2MAC, only perform well on few datasets. It means that SDSNE is efficient for multiview clustering. (3) The accuracy of SDSNE$_{\rm sc}$ on Three Sources is over 15.4\% than the best of other methods. At the same time, SDSNE is much better than other methods in other datasets and metrics, which shows that SDSNE effectively learns to obtain a good representation. In the textual and image datasets, SDSNE obtains better performance. The performance mainly depends on the quality of raw input data but does not depend on the types.

Figs.~\ref{filter}(a)-(g) shows the t-SNE~\cite{maaten2008visualizing} visualization of the raw feature and the output $H$ of SDSNE on MSRC-v1. SDSNE integrates multiple graphs into a unified graph $H$  and obtains better results than others. The t-SNE visualization and similarity graph of SDSNE show that SDSNE learns to obtain a high quality graph representation since both SDSNE$_{\rm sc}$ and SDSNE$_{\rm km}$ perform well. The $k$-means is performed well on the linear superable subspace, while SC does well in the local-closed structural data. The distribution of data points determines that they are more suitable for performing the $k$-means or SC. If their results of SDSNE are comparable, it implies the learned graph representation is of high quality as shown Fig.~\ref{filter}(h). That is the output similarity graph $H$ of MSRC-{v1} and it is a good block-diagonal matrix.
\subsection{Ablation Study}
We also explore multi-layer SDSNE. The following layer after the fist layer is given by
\begin{equation}\label{Layer2}
H_{l}^{(v)}\leftarrow H_{l-1}^{(v)}W_{l}(H_{l-1}^{(v)})^\T\,,\forall\,v\in\{1,2,\ldots,n_v\}\,.
\end{equation}
where $l>1$ is the layer index and $W_{l}$ is the parameter matrix. We add 1 to 15 layers and obtain the same results as only one layer of Eq.~\eqref{Layer}\,. Since we share $W$ between views, we obtain good results. If we use different $P^{(v)}$ from different views in Eq.~\eqref{Layer}, we obtain little bit lower results than Eq.~\eqref{Layer}. E.g, we use $P^{(1)}$ and $P^{(2)}$ at the first layer, $P^{(2)}$ and $P^{(3)}$ at the second layer, and $P^{(2)}$ and $P^{(1)}$ at the third for Three Sources dataset. SDSNE aims to learn a graph in which nodes in each connected component fully connect with each other by the same edge weight. Using the same $P^{(v)}$ rather than using cross-view diffusion obtains the target easily. Results of MSRC-v1 are shown in Table~\ref{MSRC_as}\,.
\begin{table}[tp]
  \caption{\textbf{SDSNE} performance using different layers.}
  \label{MSRC_as}
  \centering
\begin{tabular*}{0.46\textwidth}{@{\extracolsep{\fill}\quad}l|ccc}
\hline
    Method &NMI&ACC&ARI\\
\hline
Cross-view
&0.860&0.924&0.824\\
Multiple Layers
&0.872&0.933&0.845\\
SDSNE$_{\rm sc}$
&0.872&0.933&0.845\\
\hline
  \end{tabular*}
\end{table}

\section{Conclusion}

Since SDSNE aims to learn a graph in which nodes in each connected component fully connect by the same edge weight, the learned graph quality is better than other related methods. The advantages of SDSNE and the reasons why it obtains a better graph are following: (1) Its layer is based on a diffusion step of a hypergraph. SDSNE learns a graph in which nodes in each connected component fully connect by the same edge weight. (2) With co-supervision between different views, the loss function guides SDSNE in achieving the stationary state. When SDSNE achieves the stationary state, the learned graph tends to be a structure in which nodes in each connected component fully connect by the same weight.

We specifically design the Stationary Diffusion State Neural Estimation (SDSNE) approach to the stationary state. SDSNE fuses synergistically multiview structural information by a parameter-shared attentional module and learns to attain multiple graphs eventually. Using the learned graphs, we propose a structure level co-supervised learning strategy which is utilized by SDSNE to achieve the stationary state. We use the structure-level co-supervised learning strategy as the loss function which guides SDSNE in capturing consensus information. The unified consensus graph is obtained by the fusion of all learned graphs. Experiments on six real-world datasets show that SDSNE achieves state-of-the-art results for unsupervised multiview clustering.
\section*{Acknowledgements}
This work was supported by the National Natural Science Foundation of China under the Grant No.~62176108, the Natural Science Foundation of Gansu Province of China under Grant No.~20JR5RA246, and the Fundamental Research Funds for the Central Universities under the Grant No.~lzujbky-2021-ct09. The main contribution of ideas and writing was Kun Zhan whom led the author team. The four authors worked together cohesively and hardly. We thank Chang Tang for sharing datasets. We thank Chenxiao Zhan for the help of Fig.~1.
\bibliographystyle{aaai22}
\bibliography{ID_184}

\begin{thebibliography}{32}
\providecommand{\natexlab}[1]{#1}

\bibitem[{Bai et~al.(2017{\natexlab{a}})Bai, Bai, Tian, and
  Latecki}]{bai2017regularized}
Bai, S.; Bai, X.; Tian, Q.; and Latecki, L.~J. 2017{\natexlab{a}}.
\newblock Regularized diffusion process for visual retrieval.
\newblock In \emph{AAAI}, 3967--3973.

\bibitem[{Bai et~al.(2017{\natexlab{b}})Bai, Zhou, Wang, Bai, Jan~Latecki, and
  Tian}]{bai2017ensemble}
Bai, S.; Zhou, Z.; Wang, J.; Bai, X.; Jan~Latecki, L.; and Tian, Q.
  2017{\natexlab{b}}.
\newblock Ensemble diffusion for retrieval.
\newblock In \emph{ICCV}, 774--783.

\bibitem[{Blum and Mitchell(1998)}]{blum1998combining}
Blum, A.; and Mitchell, T. 1998.
\newblock Combining labeled and unlabeled data with co-training.
\newblock In \emph{COLT}, 92--100.

\bibitem[{{Chopra}, {Hadsell}, and {LeCun}(2005)}]{LeCun1467314}
{Chopra}, S.; {Hadsell}, R.; and {LeCun}, Y. 2005.
\newblock Learning a similarity metric discriminatively, with application to
  face verification.
\newblock In \emph{CVPR}, 539--546.

\bibitem[{Chung(1997)}]{chung1997spectral}
Chung, F.~R. 1997.
\newblock \emph{Spectral Graph Theory}.
\newblock American Mathematical Society.

\bibitem[{Fan et~al.(2020)Fan, Wang, Shi, Lu, Lin, and Wang}]{fan2020one2multi}
Fan, S.; Wang, X.; Shi, C.; Lu, E.; Lin, K.; and Wang, B. 2020.
\newblock One2multi graph autoencoder for multi-view graph clustering.
\newblock In \emph{WWW}, 3070--3076.

\bibitem[{Fei-Fei and Perona(2005)}]{fei2005bayesian}
Fei-Fei, L.; and Perona, P. 2005.
\newblock A bayesian hierarchical model for learning natural scene categories.
\newblock In \emph{CVPR}, volume~2, 524--531.

\bibitem[{Gao et~al.(2015)Gao, Nie, Li, and Huang}]{gao2015multi}
Gao, H.; Nie, F.; Li, X.; and Huang, H. 2015.
\newblock Multi-view subspace clustering.
\newblock In \emph{ICCV}, 4238--4246.

\bibitem[{Hadsell, Chopra, and LeCun(2006)}]{hadsell2006dimensionality}
Hadsell, R.; Chopra, S.; and LeCun, Y. 2006.
\newblock Dimensionality reduction by learning an invariant mapping.
\newblock In \emph{CVPR}, 1735--1742.

\bibitem[{Huang et~al.(2019)Huang, Zhou, Peng, Zhang, Zhu, and
  Lv}]{huang2019multi}
Huang, Z.; Zhou, J.~T.; Peng, X.; Zhang, C.; Zhu, H.; and Lv, J. 2019.
\newblock Multi-view spectral clustering network.
\newblock In \emph{IJCAI}, 2563--2569.

\bibitem[{Kipf and Welling(2017)}]{kipf2016semi}
Kipf, T.~N.; and Welling, M. 2017.
\newblock Semi-supervised classification with graph convolutional networks.
\newblock In \emph{ICLR}.

\bibitem[{Kumar and Daum{\'e}(2011)}]{kumar2011cot}
Kumar, A.; and Daum{\'e}, H. 2011.
\newblock A co-training approach for multi-view spectral clustering.
\newblock In \emph{ICML}, 393--400.

\bibitem[{Kumar, Rai, and Daume(2011)}]{kumar2011co}
Kumar, A.; Rai, P.; and Daume, H. 2011.
\newblock Co-regularized multi-view spectral clustering.
\newblock In \emph{NeurIPS}, 1413--1421.

\bibitem[{Li et~al.(2015)Li, Nie, Huang, and Huang}]{li2015large}
Li, Y.; Nie, F.; Huang, H.; and Huang, J. 2015.
\newblock Large-scale multi-view spectral clustering via bipartite graph.
\newblock In \emph{AAAI}, 2750--2756.

\bibitem[{Liu et~al.(2013)Liu, Lin, Yan, Sun, Yu, and Ma}]{6180173}
Liu, G.; Lin, Z.; Yan, S.; Sun, J.; Yu, Y.; and Ma, Y. 2013.
\newblock Robust recovery of subspace structures by low-rank representation.
\newblock \emph{TPAMI}, 35(1): 171--184.

\bibitem[{Maaten and Hinton(2008)}]{maaten2008visualizing}
Maaten, L. v.~d.; and Hinton, G. 2008.
\newblock Visualizing data using t-{SNE}.
\newblock \emph{JMLR}, 9(11): 2579--2605.

\bibitem[{Ng, Jordan, and Weiss(2002)}]{ng2002spectral}
Ng, A.~Y.; Jordan, M.~I.; and Weiss, Y. 2002.
\newblock On spectral clustering: Analysis and an algorithm.
\newblock In \emph{NeurIPS}, 849--856.

\bibitem[{Nie, Wang, and Huang(2014)}]{nie2014clustering}
Nie, F.; Wang, X.; and Huang, H. 2014.
\newblock Clustering and projected clustering with adaptive neighbors.
\newblock In \emph{KDD}, 977--986.

\bibitem[{Page et~al.(1999)Page, Brin, Motwani, and
  Winograd}]{page1999pagerank}
Page, L.; Brin, S.; Motwani, R.; and Winograd, T. 1999.
\newblock The PageRank citation ranking: Bringing order to the web.
\newblock Technical report, Stanford InfoLab.

\bibitem[{Shaham et~al.(2018)Shaham, Stanton, Li, Nadler, Basri, and
  Kluger}]{shaham2018spectralnet}
Shaham, U.; Stanton, K.; Li, H.; Nadler, B.; Basri, R.; and Kluger, Y. 2018.
\newblock Spectral{N}et: Spectral clustering using deep neural networks.
\newblock In \emph{ICLR}.

\bibitem[{Shi and Malik(2000)}]{ShijianboCVPR}
Shi, J.; and Malik, J. 2000.
\newblock Normalized cuts and image segmentation.
\newblock \emph{TPAMI}, 22(8): 888--905.

\bibitem[{Tang et~al.(2020)Tang, Liu, Zhu, Zhu, Luo, Wang, and
  Gao}]{tang2020cgd}
Tang, C.; Liu, X.; Zhu, X.; Zhu, E.; Luo, Z.; Wang, L.; and Gao, W. 2020.
\newblock {CGD}: Multi-view clustering via cross-view graph diffusion.
\newblock In \emph{AAAI}, 5924--5931.

\bibitem[{Von~Luxburg(2007)}]{von2007tutorial}
Von~Luxburg, U. 2007.
\newblock A tutorial on spectral clustering.
\newblock \emph{Statistics and Computing}, 17(4): 395--416.

\bibitem[{Wang et~al.(2012)Wang, Jiang, Wang, Zhou, and Tu}]{CVPR6248029}
Wang, B.; Jiang, J.; Wang, W.; Zhou, Z.-H.; and Tu, Z. 2012.
\newblock Unsupervised metric fusion by cross diffusion.
\newblock In \emph{CVPR}, 2997--3004.

\bibitem[{Wang et~al.(2014)Wang, Mezlini, Demir, Fiume, Tu, Brudno,
  Haibe-Kains, and Goldenberg}]{wang2014similarity}
Wang, B.; Mezlini, A.~M.; Demir, F.; Fiume, M.; Tu, Z.; Brudno, M.;
  Haibe-Kains, B.; and Goldenberg, A. 2014.
\newblock Similarity network fusion for aggregating data types on a genomic
  scale.
\newblock \emph{Nature Methods}, 11(3): 333--337.

\bibitem[{Wang, Yang, and Liu(2020)}]{8662703}
Wang, H.; Yang, Y.; and Liu, B. 2020.
\newblock {GMC}: Graph-based multi-view clustering.
\newblock \emph{TKDE}, 32(6): 1116--1129.

\bibitem[{Xia et~al.(2014)Xia, Pan, Du, and Yin}]{xia2014robust}
Xia, R.; Pan, Y.; Du, L.; and Yin, J. 2014.
\newblock Robust multi-view spectral clustering via low-rank and sparse
  decomposition.
\newblock In \emph{AAAI}, 2149--2155.

\bibitem[{Xie, Girshick, and Farhadi(2016)}]{xie2016unsupervised}
Xie, J.; Girshick, R.; and Farhadi, A. 2016.
\newblock Unsupervised deep embedding for clustering analysis.
\newblock In \emph{ICML}, 478--487.

\bibitem[{Zhan et~al.(2019)Zhan, Nie, Wang, and Yang}]{Zhan8052206}
Zhan, K.; Nie, F.; Wang, J.; and Yang, Y. 2019.
\newblock Multiview consensus graph clustering.
\newblock \emph{TIP}, 28(3): 1261--1270.

\bibitem[{Zhan et~al.(2018)Zhan, Zhang, Guan, and Wang}]{zhancyb2018}
Zhan, K.; Zhang, C.; Guan, J.; and Wang, J. 2018.
\newblock Graph learning for multiview clustering.
\newblock \emph{TCyb}, 48(10): 2887--2895.

\bibitem[{Zhang et~al.(2018)Zhang, Fu, Hu, Cao, Xie, Tao, and
  Xu}]{zhang2018generalized}
Zhang, C.; Fu, H.; Hu, Q.; Cao, X.; Xie, Y.; Tao, D.; and Xu, D. 2018.
\newblock Generalized latent multi-view subspace clustering.
\newblock \emph{TPAMI}, 42(1): 86--99.

\bibitem[{Zhou and Burges(2007)}]{zhou2007spectral}
Zhou, D.; and Burges, C.~J. 2007.
\newblock Spectral clustering and transductive learning with multiple views.
\newblock In \emph{ICML}, 1159--1166.

\end{thebibliography}
\end{document}